# A self-organizing system for urban traffic control based on predictive interval microscopic model

**Bartłomiej Płaczek**

Institute of Computer Science, University of Silesia

Będzińska 39, 41-200 Sosnowiec, Poland

E-mail: placzek.bartlomiej@gmail.com

Tel.: +48 323689793, Fax: +48 323689760

**Abstract:** This paper introduces a self-organizing traffic signal system for an urban road network. The key elements of this system are agents that control traffic signals at intersections. Each agent uses an interval microscopic traffic model to predict effects of its possible control actions in a short time horizon. The executed control action is selected on the basis of predicted delay intervals. Since the prediction results are represented by intervals, the agents can recognize and suspend those control actions, whose positive effect on the performance of traffic control is uncertain. Evaluation of the proposed traffic control system was performed in a simulation environment. The simulation experiments have shown that the proposed approach results in an improved performance, particularly for non-uniform traffic streams.

INSPEC codes: (C10) Systems and control theory, (C33) Control applications

# 1. INTRODUCTION

Self-organizing systems solve different control problems by simple local interactions among a large number of components (agents). These systems are not controlled in a top-down manner by some central unit such as an intelligent operator. Instead, they are based on internal, bottom-up processes (Gershenson and Heylighen, 2003). Artificial self-organizing systems were inspired by natural systems that are regulated by their own internal processes, e.g. swarm behaviour of social insects, fishes, and birds (Floreano and Mattiussi, 2008). From an engineering point of view, the advantages of self-organization are robustness, scalability, flexibility, adaptivity, simplicity of the components and reduced cost of deployment. Applications of the self-organization paradigm are especially effective in case of inherently decentralized control problems, such as traffic control in a road network (Khamis and Gomaa, 2014; Vasirani and Ossowski, 2011).

The task of urban traffic control is to increase capacity of a road network and decrease congestion by using traffic signals (Abdoos et al., 2013; McKenney and White, 2013). To achieve this objective effectively, traffic control algorithms take into account measured and predicted traffic data as input variables. State-of-the-art traffic control algorithms were designed to optimize traffic signals for all intersections in a road network by using a centralized adaptive approach. Global optimization of traffic signals in a road network is a NP-hard problem and the solution cannot be found in real-time due to the high computational complexity. For that reason, the centralized traffic control methods are based on adaptation of some pre-calculated signalization schedules, i.e., an optimization of signalization cycle, offset and split. (Hamilton et al., 2013). Limited performance of the adaptive methods has motivated the interest in self-organizing traffic control.

The self-organizing traffic signals use a decentralized optimization scheme, which enables global coordination of the traffic streams in a road network. In this approach, traffic signals at each intersection are controlled by an agent, which makes its own decisions independently on the basis of real-time traffic data from local measurements. It means that the agent takes into account traffic conditions that exist at road segments connected to an intersection. The decentralized, local optimization of traffic signals for particular intersections can provide efficient coordination of the traffic flows at the network level, which results in so-called green waves (Cools et al., 2013). In comparison with adaptive traffic control algorithms, the self-organizing strategy is more flexible with respect to local demands and more robust to variations in the traffic flows. The self-organizing traffic control system can respond to actual real-time traffic conditions without using any pre-determined signalization schedules that are based on average traffic characteristics, e.g., average velocity (Lämmer and Helbing, 2008).

This paper introduces a method, which improves performance of the self-organized traffic control system. The considered system is composed of agents that control traffic signals at intersections. Each agent detects incoming traffic, takes decisions, and executes control actions autonomously for its intersection. The control decision determines traffic streams that will get a green signal. According to the proposed method, a novel interval microscopic traffic model is used by the agents to predict effects of their control actions in a short time horizon. It allows delays of individual vehicles to be evaluated in terms of intervals. The higher control performance is obtained by taking into account uncertainty of the control decisions. Consecutive decisions are made by the agents in constant time steps on the basis of the predicted delay intervals. Since the prediction results are represented by intervals, the agents can recognize and suspend those control actions, whose positive effect on the performance of traffic control is uncertain. Sources of this uncertainty are associated with various driver behaviours and vehicle parameters that correspond to a range of possible free flow velocities observed in real-world (non-uniform) traffic.

The paper is organized as follows. Section 2 includes review of related research and describes main contribution of this paper. Section 3 presents a decision algorithm for traffic control agents. Details of the proposed self-organized traffic control approach are discussed in Section 4. Section 5 describes experimental setting and presents results of the simulation experiments that were conducted in order to evaluate the performance of the self-organizing traffic control. In this study, the performance of the proposed method was compared against results obtained for state-of-the-art algorithms. Finally, conclusions are given and future research directions are outlined in Section 6.

## 2. RELATED WORKS AND CONTRIBUTION

In the literature several self-organizing traffic control methods have been proposed. Some of the earliest methods were based on neural networks (Nakatsuji and Kaku, 1991) and fuzzy logic (Chiu and Chand, 1993). In (Sekiyama et al., 2001) a model of nonlinear oscillators with the nearest neighbourhood coupling was used to formulate the self-organizing traffic control strategy. Another early work along these lines (Wei et al., 2005) employs macroscopic two-dimensional cellular automata model of an urban traffic control system, in which each intersection is regarded as a cell and the flow pressure is treated as a state of the cell.

Main advantages of the self-organizing traffic control were discussed in (Gershenson, 2005). It was demonstrated that traffic signals are able to self-organize and adapt to changing traffic conditions by using simple rules without direct communication among intersections. The simple self-organizing traffic lights algorithm (SOTL) proposed in (Gershenson, 2005) gives preference to vehicles that have been waiting longer, and to larger groups of vehicles (platoons). According to that approach, platoons affect the behaviour of traffic lights, prompting them to turn green. A self-organized coordination of the traffic lights is achieved by probabilistic formation of the vehicle platoons. The SOTL algorithm was further applied to to control traffic signals in a model of hexagonal road network with complex intersections (Gershenson and Rosenblueth, 2012) as well as in a simulated real-world arterial road (Cools et al., 2013). In order to enable cooperation of the SOTL algorithm with the existing low-cost vehicle detection technology, a history-based self-organizing traffic control was proposed in (Burguillo-Rial et al., 2009) as an extension of the above works. de Gier et al. (2011) have generalized the SOTL algorithm to handle intersections with multiple signal phases and to take into acconut the situation when vehicles cannot exit the intersection due to congestion in downstream links. In (Zhang et al., 2013) SOTL was compared against SCATS (Sydney Coordinated Adaptive Traffic System), i.e., a traffic signal system, which at present controls the traffic signals in numerous cities around the world. Results of those studies show that the self-organizing traffic control method considerably improves the performance of the conventional methods and increases the network capacity.

Self-organizing oscillatory changes of passing direction observed in pedestrian counterflows at bottlenecks were an inspiration for the control strategy reported in (Helbing et al., 2005). This method assumes a control of traffic lights based on priorities that correspond to "pressures" induced by vehicles waiting to be served at an intercection. For the purpose of this strategy, a macroscopic fluid-dynamic model was used to make short-term traffic flow predictions. Performance of this self-organizing traffic control was compared to that of the existing state-of-the-art adaptive control in a real-world road network (Lämmer and Helbing, 2008). Results of these experiments revealed the higher performance of the self-organizing approach. Szklarski (2010) has shown that this approach always gives better results than any solution based on fixed cycles with green waves.

In recent years there has been considerable interest in the developement of self-organizing traffic control systems. Main reason of this interest is that the improved self-organized coordination of traffic signals can have a noticeable effect on the quality of life in cities (Gershenson, 2013). Among the latest studies, Wang and Liu (2013) have proposed a self-organized control strategy in which each intersection optimizes the traffic signals by taking into acount not only its own benefit, but also the interactions between neighbouring intersections. Cesme and Furth (2013) have introduced rules that can be added to the self-organizing control logic for managing queues in arterial traffic during periods of oversaturation. In (Suzuki et al., 2013) it was suggested that the self-organizing traffic control system exhibits chaotic behaviour and therefore it can be analysed by using the Ising model of ferromagnetism in statistical mechanics. According to that appraoch, the states of traffic signals at intersections are represented by atomic spins on a two-dimensional lattice.

Main limitations of the above self-organizing traffic control methods result from the use of macroscopic or mesoscopic traffic models for prediction and decision-making purposes. Such models evaluate the total effect of traffic signals on delay for all or a group of vehicles in a given road section. They assume uniform traffic streams and ignore individual vehicle parameters. It means that the same (average) parameters are used to describe motion of each vehicle. For real-world traffic, the motion parameters differ significantly depending on vehicle type and driver behaviour. Such non-uniform traffic scenarios cannot be taken into account while using the existing methods.

By contrast, the new approach introduced in this paper is based on a microscopic traffic model, which uses intervals to describe possible range of vehicle parameters for the non-uniform (heterogeneous) traffic streams. Thus, the traffic model takes into account different vehicle types and driver behaviours. The proposed method enables prediction of the effect of signal control actions on delays for individual vehicles. It allows the traffic control agents to utilize microscopic data (e.g., vehicle positions, velocities, classes, etc.) that are available in video-detection systems (Płaczek and Staniek, 2007; Pamuła, 2012), vehicular sensor networks (Bernaś, 2012; Płaczek, 2012) and other modern traffic monitoring platforms. Since the results of prediction are represented by intervals, the uncertainty of control decisions can be analysed before execution of a control action. Therefore the agents can recognize and suspend those control actions, whose positive effect on the performance of traffic control is uncertain.

## 3. DECISION ALGORITM FOR TRAFFIC CONTROL AGENTS

This section presents the main features and assumptions of the decision algorithm for control agents in a self-organizing traffic signal system. The objective of the considered system is to minimise delay times in a road network by controlling traffic signals at intersections. Components of the system are agents that detect incoming traffic, take decisions, and execute control actions independently for each intersection in the network.

Pseudo-code of a decision algorithm for the traffic control agents is presented in Algorithm 1. Input data of the algorithm consist of parameters that describe incoming traffic streams. These data are collected from vehicle detectors at road segments that enter the intersection. Output of the algorithm is a control decision that determines which traffic stream should get a green signal. The consecutive control decisions are made in constant time steps.

In order to make control decision the agent uses a traffic model for approximation of the current traffic state as well as for prediction of its future evolution. On the basis of the prediction results a cost function $C(\alpha)$ is evaluated for all possible control actions $\alpha = 1, \ldots, m$. The term "control action" refers to the operation of providing green signal for a selected traffic stream (or streams). Thus, the maximum number of possible control actions is equal to the number of incoming traffic streams. Actual number of possible control actions $m$ can be lower than the maximum if some of the traffic streams are not in conflict and thus can get a green signal at the same time. Since the control objective is to minimize delay time, the cost function is usually defined as total delay of vehicles.

Control decision indicates the control action that will be executed by the agent at current time step. If this action changes the traffic stream receiving green signal then a setup time (intergreen period) $\tau$ has to be introduced due to safety requirements. During setup time the decision making procedure is skipped because the service cannot be switched to another traffic stream.

In general, according to the discussed algorithm each agent executes an optimal control action, which is associated with the lowest cost $C(\alpha)$. However, the decision algorithm has also to assure that all traffic streams will be served in a critical time window $T_{crit}$. Therefore the costs are taken into account after checking the time window constraint for all possible control actions. If the time window $T(\alpha)$, predicted for control action $\alpha$, is equal to or longer than the critical time $T_{crit}$, then the control action $\alpha$ is executed immediately and the costs are ignored.

**Algorithm 1.** A general decision algorithm for traffic control agents

```
1   for each time step do
2   begin
3      estimate current traffic state
4      if not setup time then
5         for each control action α = 1 … m do
6         begin
7            predict C(α) and T(α)
8            if T(α) >= T_crit then
9            begin
10              execute control action α
11              go to 16
12           end
13        end
14        select optimal control action α*
15        execute control action α*
16  end
```

Detailed definitions of the above discussed operations depend mainly on the traffic model used by the control agents for prediction purposes. The strategy developed by Lämmer and Helbing (2008), discussed in Sect. 5.1.2, as well as the proposed approach, which is introduced in the next section, can be considered as special cases of the general decision algorithm presented here.

## 4. PROPOSED APPROACH

In this section details are provided regarding the proposed self-organizing traffic signal control for an urban road network. The main novelty of the introduced method is that the control decisions are based on predictions obtained from interval microscopic traffic model. It allows the delays with their uncertainties to be analysed for individual vehicles in heterogeneous traffic streams.

**4.1. Interval microscopic model**

The interval microscopic model of traffic stream combines cellular automata approach (Maerivoet and De Moor, 2005) with interval-based representation of vehicle parameters. Due to the microscopic level of details, the model can directly map parameters of individual vehicles. It means that the results of vehicle detection can be directly taken into account in making control decisions. Cellular automata are used to simulate future evolution of traffic streams. The interval-based representation allows the model to take into account the uncertainty associated with vehicle parameters and drivers behaviour.

According to this method, traffic streams are modelled in discrete time and space. A traffic lane is divided into segments of equal length that are represented by cells in the cellular automaton. Traffic streams at an intersection correspond to ordered sets of vehicles. Each vehicle ($i$) in a traffic lane is described by its current position $X_i(t)$ (occupied cell) and velocity $V_i(t)$ (in cells per time step). The maximum velocity of vehicles is defined by parameter $V_{max}$. Velocities and positions of all vehicles are computed simultaneously in discrete time steps of one second using a transition rule of the cellular automata.

A characteristic feature of this model is the application of intervals for describing vehicle position. The following notation will be used to express the position of vehicle $i$ at time step $t$: $X_i(t) = [x_i^-(t), x_i^+(t)]$. Velocity of a vehicle is defined as an ordered pair $V_i(t) = (v_i^-(t), v_i^+(t))$, where the first entry corresponds to position $x_i^-(t)$ and the second entry to position $x_i^+(t)$. Note that the relation $v_i^-(t) \leq v_i^+(t)$ does not need to be satisfied. The above position interval and velocity pair are updated at each time step $t$ according to the cellular automata rule. In case of lower endpoint of the position interval, the update operation is defined by formulas:

$$v_i^-(t) = \min\{v_i^-(t-1)+1, g_i^-(t), v_{max}^-\}, \qquad (1)$$

$$x_i^-(t+1) = x_i^-(t) + v_i^-(t), \qquad (2)$$

where $g_i^-(t)$ denotes the number of empty cells in front of vehicle $i$ at time step $t$. Note that in Eq. 2 it is implicitly assumed that the velocity $v_i^-(t)$ is expressed in cells per one time step (1 second). Upper endpoint of the position interval $x_i^+(t)$ and the related velocity $v_i^+(t)$ are calculated in a similar way. The maximum velocity interval $V_{max} = [v_{max}^-, v_{max}^+]$ should be interpreted as an enclosure of possible free flow velocities of vehicles in a heterogeneous traffic stream.

In this study a single rule of cellular automata is applied with different values of the $v_{max}$ parameter to calculate the endpoints of position intervals. However, this approach can be extended by using two different rules of cellular automata representing two extreme cases of vehicle movement in a traffic stream. In such a model, one cellular automata rule would be used to calculate velocity $v_i^-(t)$ and position $x_i^-(t)$, while another rule would apply to the computation of $v_i^+(t)$ and $x_i^+(t)$. The method may be easily adapted to the specific needs since the appropriate rules can be selected among those available in the rich literature on cellular automata traffic models (Maerivoet and De Moor, 2005). Details of the possible model extensions are discussed in the previous work of the author (Płaczek, 2013).

**4.2. Simulation-based prediction**

In the proposed traffic control system, the traffic model is applied to a simulation-based prediction of vehicle delays. According to the concept of simulation-based prediction, the traffic model has to be synchronised with real time and adjusted to traffic data collected from vehicle detectors. To maintain consistency between simulated and real traffic, at each time step, the model is appropriately adjusted by taking into account the current traffic situation (real time simulation). After adjustment, the traffic model determines initial conditions for faster than real time simulation. During this simulation the vehicle delays are predicted. The simulation-based prediction technique enables rapid evaluation of alternate courses of action in order to aid the control agents in decision making processes.

Due to low computational complexity, the interval microscopic model enables high speed traffic simulation. Thus, it can be used to estimate current traffic state (real time simulation) and to predict delay of vehicles at an intersection for all available control actions (faster than real time simulation). The predicted delays are taken into account by agents for evaluation of the cost function $C(\alpha)$ during making control decisions.

Estimation of current traffic state is based on both the real traffic data acquired from vehicle detectors and the results of real time simulation. During the real time simulation traffic model is used to estimate current positions and velocities of vehicles that cannot be measured by the detectors. In order to facilitate real time simulation, the detector data are mapped into the traffic model. This operation includes generation and removal of vehicles, updating their positions, and adjusting the maximum velocity parameters. The real time simulation has to take into account also actual status of traffic signals.

Results of the real time simulation (i.e., data on individual vehicles approaching an intersection) are further used to determine initial configurations for faster than real time simulation. The task of the faster than real time simulation is to predict values of the cost function for all possible control actions. For traffic stream, which currently receives green signal, the prediction horizon is equal to the minimum green time. In case of a traffic stream having red signal the prediction horizon is extended by the duration of intergreen period. Using the interval microscopic model, the cost associated with control action $\alpha$ is calculated as an interval of delay $C(\alpha) = [c^-(\alpha), c^+(\alpha)]$. The following formulas apply for determining the endpoints of this interval:

$$c^-(\alpha) = \min\left\{\sum_t\sum_i w_i^-(t), \sum_t\sum_i w_i^+(t)\right\}, \quad (3)$$

$$c^+(\alpha) = \max\left\{\sum_t\sum_i w_i^-(t), \sum_t\sum_i w_i^+(t)\right\}, \quad (4)$$

$$w_i^-(t) = \begin{cases} 1, & v_i^-(t) = 0, \\ 0, & \text{else.} \end{cases} \quad (5)$$

$$w_i^+(t) = \begin{cases} 1, & v_i^+(t) = 0, \\ 0, & \text{else.} \end{cases} \quad (6)$$

Simulation examples for two different initial configurations of the interval microscopic model are illustrated in Fig. 1. It was assumed that the analysed road section consists of 12 cells. Stop line is located in cell 11 and red signal is displayed for the vehicles during all simulation period. The prediction horizon is 6 time steps long. It was assumed that initial positions of three approaching vehicles at time step 0 are known (determined on the basis of detection results). In example a) the vehicles at time step 0 are detected in cells 1, 3, and 10. For example b) the three vehicles occupy cells 0, 4, and 8. Maximum velocity is given by interval $V_{\max} = [1, 2]$ (in cells per time step). Shaded regions in the charts (Fig. 1 a, b) correspond to predicted position intervals. Fig 1 c) presents the delay intervals calculated for simulation presented in Fig. 1 a). Result of the delay prediction in this example is the interval $C(\alpha) = [8, 14]$. For the example in Fig. 1 b) the result is $C(\alpha) = [7, 13]$, as shown in Fig 1 d).

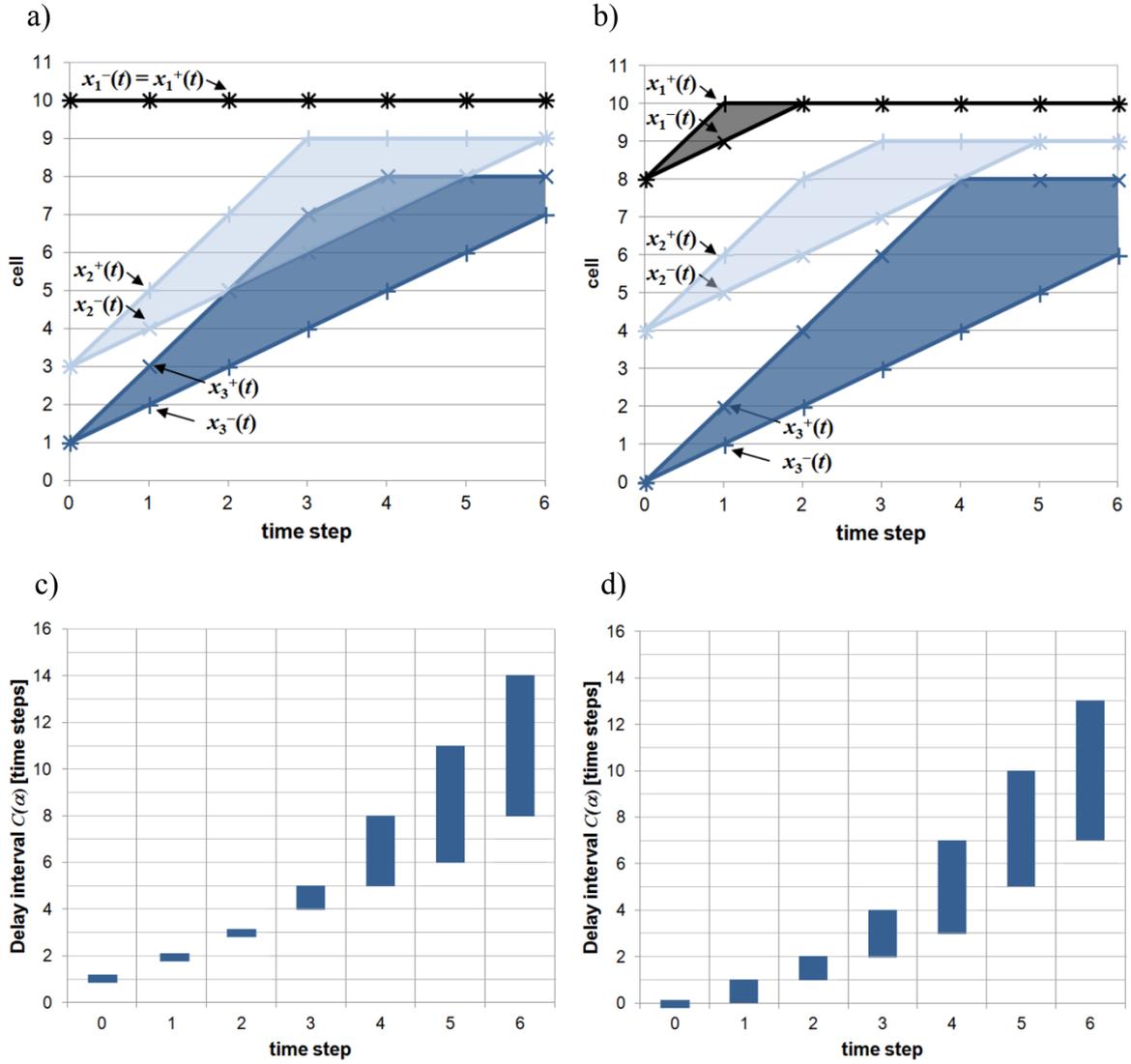

**Fig. 1.** Traffic stream simulation in interval microscopic model: a, b) vehicle position intervals, c, d) delay intervals.

### 4.3. Selection of control action

Control decision, i.e., the selection of the most preferable (optimal) control action $\alpha^*$ is made on the basis of the predicted delay (cost). Intuitively, the agent should seek for a minimum element in $\{C(\alpha)\}$, $\alpha = 1, \ldots, m$. However, this operation requires a definition of an order relation for intervals. In the presented approach two different definitions of such relation are used.

The first definition of the order relation between two intervals $A$ and $B$ was taken from the Moore's work (Moore, 1979). This basic definition is formulated as follows:

$$A < B \quad \text{iff} \quad a^+ < b^-. \tag{7}$$

Moore also defined the equality of two intervals as

$$A = B \quad \text{iff} \quad a^- = b^- \quad \text{and} \quad a^+ = b^+. \tag{8}$$

Above definition of the order relation "<" is applicable only for non-overlapping intervals. Therefore, the second definition of relation "$\prec$" has been adopted for the case of overlapping intervals (Ishibuchi and Tanaka, 1990):

$$A \prec B \quad \text{iff} \quad A \prec_= B \quad \text{and} \quad A \neq B, \tag{9}$$
$$A \prec_= B \quad \text{iff} \quad a^- \leq b^- \quad \text{and} \quad a^+ \leq b^+. \tag{10}$$

It should be also noted here that the relations "<" and "$\prec$" satisfy the following property, which was used to develop a procedure for selection of an optimal control action:

$$A < B \quad \text{and} \quad C \prec A \quad \Rightarrow \quad C < B. \tag{11}$$

If the comparison is made for intervals representing costs of control actions, then the relation $C(1) < C(2)$ indicates that the control action 1 is certainly better than the control action 2. On the other hand, the relation $C(1) \prec C(2)$ tells that the first control action is probably more cost effective than the second one, thus the order relation "$\prec$" involves some level of uncertainty.

Each change of the currently executed control action $\alpha^C$ requires an additional setup time before switching the green signal. During this time, the approaching vehicles have to wait at the intersection. Therefore, the proposed procedure (Algorithm 2), selects a new control action $\alpha$ only if its cost is certainly lower than the cost predicted for the current control action, i.e., $C(\alpha) < C(\alpha^C)$. If there is more than one control action, which satisfies the above condition then the final selection is made using the uncertain order relation "$\prec$".

According to the proposed procedure, the decision about changing the current control action $\alpha^C$ is made only if it is certain. In case of uncertainty, the decision is to continue the execution of the control action $\alpha^C$. Symbol $\alpha^*$ was used in the pseudo-code to denote the selected control action, which will be executed by the agent at an intersection. Illustrative examples of selecting a control action are shown in Fig. 2. In the left example, the execution of current control action (1) is continued because it is uncertain if the costs of control actions 2 or 3 will be lower than the cost of 1. For the two remaining examples, the control action 3 is selected since its predicted cost $C(3)$ is certainly lower than $C(1)$ and possibly lower than $C(2)$.

**Algorithm 2.** Procedure for selection of control action.
```
1   α* := α^C
2   α := 0
3   while α < m and α* = α^C do
4   begin
5       α := α + 1
6       if C(α) < C(α^C) then α* := α
7   end
8   while α < m do
9   begin
10      α := α + 1
11      if C(α) ≺ C(α*) then α* := α
12  end
```

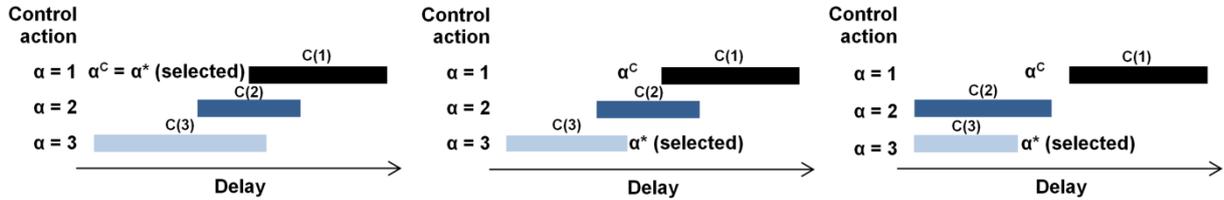

**Fig. 2.** Selecting control action on the basis of cost intervals.

## 5. EXPERIMENTS

The main goal of the experimental evaluation is to compare performance of the proposed approach with that of the existing self-organizing traffic control methods. The comparison was made for the non-uniform urban traffic, where the vehicles have different free-flow velocities. Average delay per vehicle was used in this study as the performance measure. The experiments were conducted in a simulation environment that utilizes a model of urban road network based on stochastic cellular automata.

In order to demonstrate the benefits of the proposed approach, a comparative analysis was performed taking into account previous self-organizing traffic control methods that were introduced by Helbing et al. (2005) and Gershenson (2005). Different versions of the agent decision algorithm were examined to verify the effect of using the simulation-based prediction technique and the interval microscopic model.

Moreover, the influence was investigated of detailed (microscopic) input data on the performance of self-organizing traffic control. To this end, an application of two different vehicle detection systems was considered, i.e., road-side vehicle detectors and a vehicular sensor network. For the road-side detectors it was assumed that vehicles can be detected (i.e., their positions can be determined) only when passing intersections or entering the road network. In the case of vehicular sensor network the complete information on vehicles positions is available at each time step of the simulation.

### 5.1. Compared algorithms

Hereinafter the acronyms SOTL, SOC, SOC_2, SOC_M, and SOC_2M will be used to refer the considered agent decision algorithms for the self-organizing traffic control system. Table 1 shows which elements of the new approach (introduced in Sect. 4) are combined with the state-of-the-art methods for the examined algorithms. Details of these algorithms are discussed in the following subsections.

**Tab. 1.** Compared agent decision algorithms.

| Algorithm | Traffic model | Cost prediction | Selection of control action |
|---|---|---|---|
| SOTL | State-of-the-art | None | State-of-the-art |
| SOC | State-of-the-art | State-of-the-art | State-of-the-art |
| SOC_2 | Interval microscopic model | Simulation-based | State-of-the-art |
| SOC_M | Interval microscopic model | State-of-the-art | Interval ordering |
| SOC_2M | Interval microscopic model | Simulation-based | Interval ordering |

### 5.1.1. SOTL

The SOTL algorithm (originally named SOTL-platoon in (Gershenson, 2005)) uses a very simple traffic model which takes into account current counts of vehicles approaching particular traffic signals at an intersection. Each traffic signal (*i*) has a counter $\kappa_i$ which is set to zero when the signal turns red and then it is incremented at each time step by the number of vehicles approaching this red signal. The vehicles are counted at a distance of 80 m from the red signal. When $\kappa_i$ reaches a threshold $\theta = 50$, the control action at the intersection is changed, i.e., a setup time is introduced and after that the signal *i* turns green.

In order to prevent the traffic signals from switching too frequently, the following minimum green time constraint is used: a green signal *i* will not be changed to red if $\varphi_i < \varphi_{min}$, where $\varphi_i$ is the time since the signal *i* turned green, $\varphi_{min} = 5$ s.

Another constraint in the SOTL algorithm was defined to regulate the size of platoons. According to this constraint, a green signal cannot be changed to red if the number of vehicles approaching this signal is between 1 and $\mu = 3$. The vehicles are counted at a distance of 25 m from the green signal. On the one hand, this condition keeps the crossing platoons together, but on the other hand, it allows for dividing the large platoons that would excessively block the traffic flow of intersecting streets. Pseudo-code of the SOTL algorithm can be found in (Cools et al., 2013).

### 5.1.2. SOC

The second considered algorithm for the self-organising traffic control system (SOC) was proposed in (Helbing et al., 2005). In this algorithm, the cost associated with a control action *α* is calculated (predicted) as a total increase of the expected delay, using the following formula:

$$C(\alpha) = N_\alpha (\tau_\alpha + G_\alpha) + \Delta w_\alpha, \qquad (12)$$

where: $N_\alpha$ denotes the number of vehicles that are expected to wait at the intersection during execution of the control action *α*, $\tau_\alpha + G_\alpha$ is the time period necessary to finish the execution of control action *α*, which consists of the remaining setup time $\tau_\alpha$ and the green time $G_\alpha$ required for vehicles to leave the intersection, $\Delta w_\alpha$ reflects the extra delay associated with the setup for switching back later to the current control action $\alpha^C$ (note that $\Delta w_\alpha = 0$ if $\alpha = \alpha^C$).

The extra delay $\Delta w_\alpha$ can be interpreted as the additional cost of terminating the current control action $\alpha^C$ and switching to $\alpha$.

According to the general decision algorithm for traffic control agents (Algorithm 1), to make a control decision, the traffic control agents have to predict also the time windows $T(\alpha)$. This prediction have to be done for all available control actions, excluding the current control action ($T(\alpha^C) = 0$). Using the approach proposed in (Helbing et al., 2005)., the time window is defined as the time interval between two successive executions of the same control action $\alpha$. Thus, the time window is calculated as follows:

$$T(\alpha) = r_\alpha + \tau_\alpha + G_\alpha, \qquad (13)$$

where $r_\alpha$ denotes the preciding red time in which the control action $\alpha$ was not executed.

In order to predict the green time $G_\alpha$, which allows detected vehicles to exit the intersection, the queuing process at the intersection is modelled as a hybrid dynamical system (Lämmer et al., 2008). This method assumes a constant velocity of vehicles in the free-flow traffic as well as a constant saturation flow. Thus, numeric (average) values of the free-flow velocity and the saturation flow have to be estimated to make the prediction. More detailed information about this algorithm can be found in (Helbing et al., 2005; Lämmer and Helbing, 2008).

5.1.3. SOC_2

SOC_2 algorithm uses the interval microscopic traffic model to predict the costs of particular control actions. This algorithm applies the simulation-based prediction method, which was introduced in Sect. 4.2. The predicted costs correspond to delays of vehicles at an intersection and are represented by intervals. In order to select the control action which will be executed, the SOC_2 algorithm finds a minimum element in a set of values that correspond to the centers of the cost intervals. Thus according to this algorithm, a control agent converts the cost intervals into numeric values before making the decision.

5.1.4. SOC_M

In SOC_M algorithm, the interval microscopic model is used to predict the green time required for vehicles to leave the intersection $G_\alpha$ and the number of waiting vehicles $N_\alpha$ (both are intervals in this case). The cost of control action, i.e., the delay, is calculated according to Eq. (12). However, these calculations are performed with application of the interval arithmetic because the simulation results ($G_\alpha$, $N_\alpha$) are processed in the form of intervals. Finally, the evaluated cost intervals are analyzed to make the control decision. This decision is made using the procedure introduced in Sect. 4.3.

5.1.5. SOC_2M

The last algorithm taken into consideration (SOC_2M) includes all the new elements of the proposed approach that were discussed in Sect. 4. According to this algorithm, the delays (costs of control actions) are evaluated directly during traffic simulation in the interval microscopic model. The cost intervals are not converted into numeric values, thus selection of the control action is performed using the interval ordering approach.

**5.2. Road network model**

The simulation experiments were conducted using an extended version of the urban traffic model based on stochastic cellular automata (BBSS), which was originally presented in (Brockfeld et al., 2001). Topology of the simulated network is a square lattice of 8 two-directional roads with 16 signalised intersections. Links between intersections consists of 40 cells that correspond to the distance of 300 m. At each intersection there are two alternative control actions: the green signal can be given to vehicles coming from north and south or to those that are coming from west and east. A snapshot of the simulation scenario is shown in Fig. 3.

The urban road network in BBSS model has square lattice geometry. All streets are unidirectional. The dynamics of vehicles on the streets is simulated by using stochastic cellular automaton with two parameters: maximum velocity $v_{max}$ and randomization parameter $p$. At each discrete time step the state of cellular automaton is updated according to a rule which includes four steps (acceleration, braking due to other vehicles or traffic light, randomization of vehicle velocity, and vehicle movement). Detailed definitions of these steps can be found in (Brockfeld et al., 2001).

A modification of the original BBSS model was introduced to take into account bidirectional streets and two classes of vehicles that differ in their free-flow velocities.

Maximal velocity $v_{max}$ in the BBSS model was set to 2 cells per time step. Values of the probabilistic randomization parameter $p$ (so-called braking probability) are: 0.2 for fast vehicles and 0.8 for slow vehicles. Figure 4 shows the obtained free-flow velocity distributions. The average values of these distributions are: 1.2 cells per time step (32.4 km/h) for slow vehicles and 1.8 cells per time step (48.6 km/h) for fast vehicles (the simulation time step is one second).

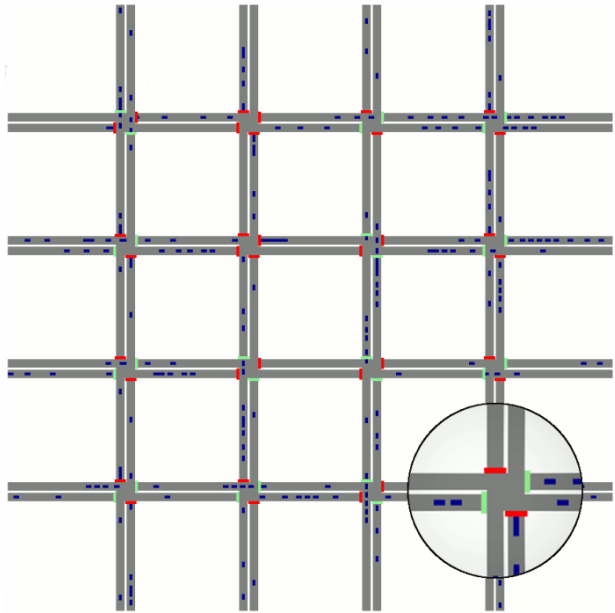

**Fig. 3.** Snapshot of the simulation scenario.

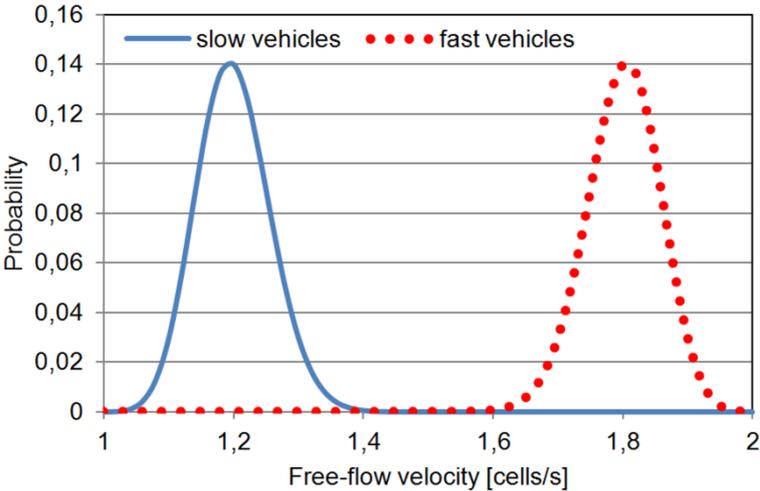

**Fig. 4.** Distribution of free-flow velocity for slow and fast vehicles.

Input parameters $q$ and $f$ are used in the network model to determine volume of the traffic flow and percentage of the slow vehicles respectively. The flow volume $q$ is expressed in vehicles per hour (vehs/h). This parameter refers to all traffic streams entering the road network - the vehicles are generated at each network entry point with the same intensity, according to the parameter $q$. The output of the model is the computed average delay per vehicle.

The self-organizing traffic control was simulated assuming that the setup times $\tau_\alpha$ are equal to 5 s and the critical time window $T_{crit}$ is 120 s. The control decisions are made by agents in time steps of 1 s. Depending on the decision algorithm, the control agents use different models for the purpose of current traffic state estimation and costs prediction (see Tab. 1). In case of the algorithms SOC_2, SOC_M, and SOC_2M the traffic streams were mapped using the interval microscopic model. In this model, the interval of maximal velocity $V_{max}$ was set to [1, 2] (in cells per time step). It allows us to take into consideration both the slow and the fast vehicles. For the state-of-the-art algorithm (SOC) the traffic model takes into account a constant free-flow velocity, which was estimated as an average over the two vehicle classes ($V = 1.5$ cells/s).

### 5.3. Experimental results

The simulation experiments were performed to evaluate performance of the agent decision algorithms in self-organizing traffic control. During experiments, both the traffic flow volume and the percentage of slow vehicles were changed in a wide range. The collected results include average delays of vehicles for various traffic conditions.

As it was already mentioned, two different simulation scenarios were based on different vehicle detection systems. In the first scenario road-side vehicle detectors (RVD) are simulated that provide information on vehicle position only for the network entrances and for the intersections. The second scenario assumes that the road traffic is monitored by a vehicular sensor network (VSN), thus the information on vehicle position is available for all road sections. For each analyzed case (given scenario, flow volume, and percentage of slow vehicles) 100 simulation runs were executed. The time period of each simulation run was 3 hours.

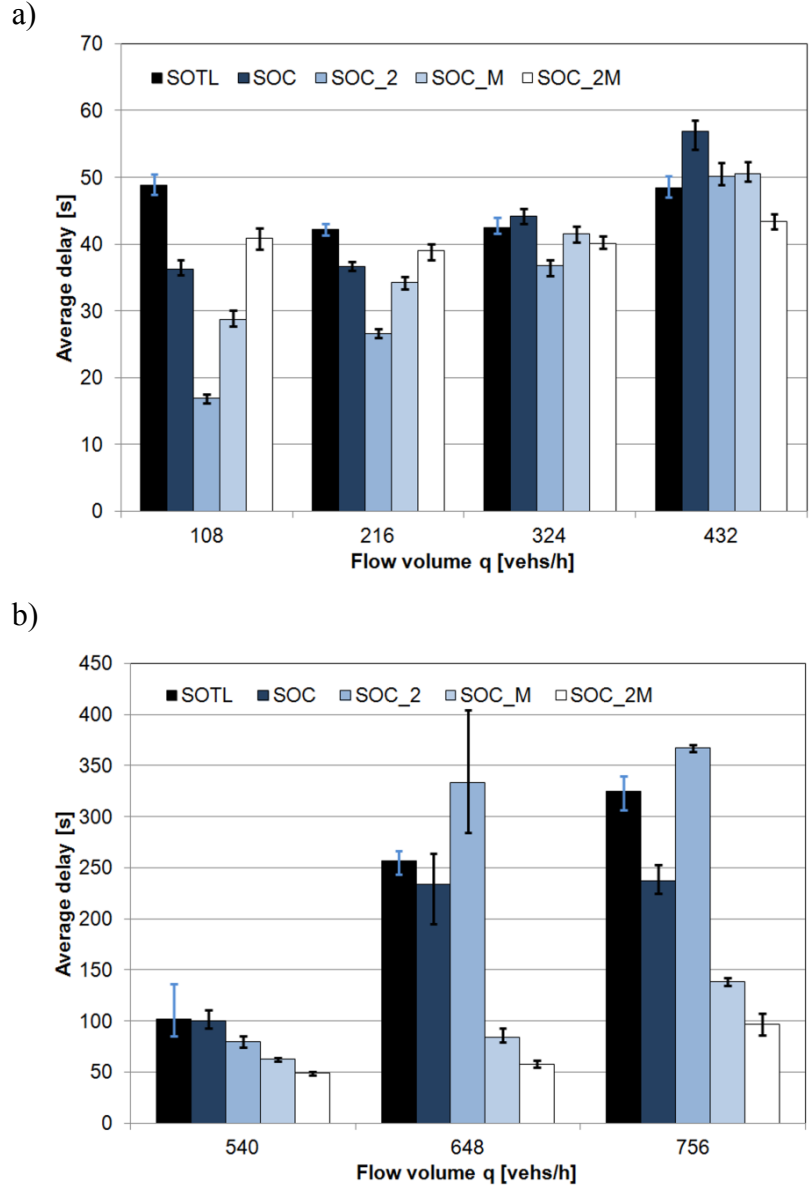

**Fig. 5.** Average delay vs. traffic flow volume (RVD scenario, $f = 20\%$): a) low and medium flow volumes, b) high flow volumes.

Results of the simulation for RVD scenario are presented in Figs. 5 and 6. The average delays in Fig. 5 were obtained assuming the 20% fraction of slow vehicles. These results show that the proposed approach outperforms the state-of-the art algorithms (SOTL and SOC). In case of high flow volumes ($q \geq 432$ vehs/h), the lowest vehicle delay was obtained by the SOC_2M algorithm. The most significant decrease in average delay (over 50%) was observed for the highest flow volume. It means that the introduced method increases capacity of the road network.

For the flow volumes below 400 vehs/h the highest performance of the self-organizing traffic control was achieved by the SOC_2 algorithm. However, this algorithm improves the control performance only for low traffic volumes. The better results of SOC_2 in comparison with SOC_M and SOC_2M for low flow volumes can be explained by the fact that SOC_2 does not take into account the decision uncertainty, it executes each decision even if the decision is uncertain. Therefore, SOC_2 can react faster to the dynamic changes of the traffic situation at low flow volumes by switching the traffic signals immediately. In case of higher traffic volumes, the frequent switching of traffic signals results in significantly increased delays, since each change of the signals at an intersection requires additional setup time to be introduced. As it can be observed in Fig. 5 for medium and high traffic volumes, the algorithms SOC_M and SOC_2M (that eliminate the uncertain decisions by using the proposed interval ordering method) allow the vehicle delays to be significantly reduced.

According to the presented results, the performance of the state-of-the-art methods (SOTL and SOC) can be improved by using one of the proposed algorithms for each of the analysed traffic conditions. It should be noted that SOTL is the simplest algorithm, which does not involve prediction and selects the control actions by taking into account only the current counts of vehicles approaching an intersection. The other considered algorithms perform a proactive optimization, i.e., take the control decisions based on the predicted effects of particular control actions. For the SOC algorithm, the prediction is less accurate than for the proposed methods, as it is obtained by using the macroscopic model, which ignores the non-uniform character of traffic streams.

Figure 6 compares the average delay registered for different percentages of the slow vehicles, with constant flow volume $q = 540$ vehs/h. It is evident from these results that the proposed algorithms improve the performance of the self-organizing traffic control for all analyzed conditions. The improvement in performance increases with the percentage of slow vehicles. The best results were obtained for SOC_2M. Interestingly, this algorithm provides stable and low average delay, which only slightly varies with respect to the changes in the percentage of slow vehicles. Similarly, in case of SOC_M algorithm the average delay does not increase significantly with the percentage of slow vehicles. This effect is a consequence of using the interval microscopic model together with the interval ordering method for selection of control action. It allows the control agents to take into account the possible range of the free-flow vehicle velocities when making control decisions. Therefore, the control agents can select the control actions that are effective for both the slow and the fast vehicles.

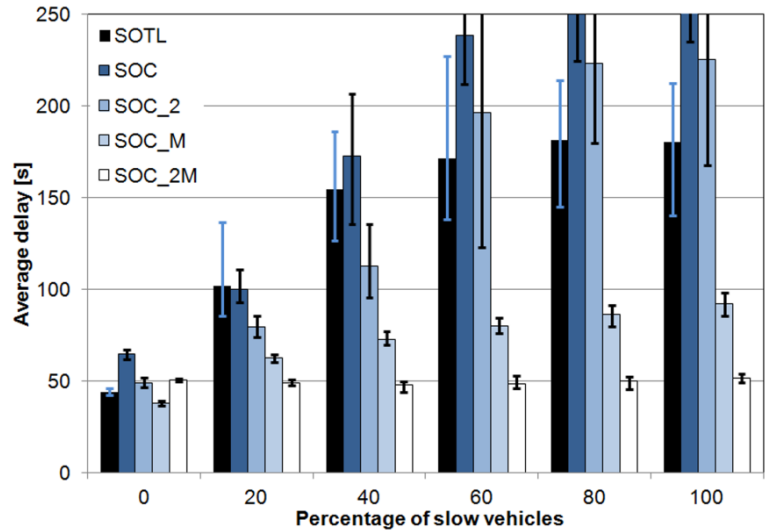

**Fig. 6.** Average delay vs. percentage of slow vehicles (RVD scenario, $q = 540$ vehs/h).

The charts in Figs. 7 and 8 show the experimental results obtained for the VSN-based vehicle detection scenario. In this scenario, the proposed algorithms also allow the self-organizing traffic control to reach higher performance. When analyzing the average delays for various flow volumes (Fig. 6), it can be observed that, as in the RVD scenario, the SOC_2 algorithm provides the lowest delay for low-traffic conditions. However, in case of high traffic volumes the best results were obtained when using SOC_M algorithm for making control decisions. It should be also noted that the SOC_2M algorithm allows the vehicle delays to be at the low level for all analysed traffic conditions. The percentage of slow vehicles in this experiment was 20%. The SOTL algorithm was not considered in the VSN-based detection scenario as it cannot utilize the additional detailed information on vehicles positions.

Dependency between average delay and the percentage of slow vehicles for the VSN scenario is shown in Fig. 7. These results were registered with the simulated traffic flow volume of 540 vehs/h. For all analysed percentages, the proposed algorithms improve the performance of the self-organizing traffic control. The observed average delays are very similar for all introduced algorithms and do not change significantly with the percentage of the slow vehicles. A distinct increase in the average delay was encountered between $f = 0\%$ and $f = 20\%$. For higher percentages of slow vehicles, the delay remains at almost constant level. The reason underlying these results arises from the fact that a low fraction of slow vehicles can significantly influence the flow of a large number of fast vehicles. For instance, one slow vehicle may reduce velocity of a large platoon of fast vehicles and the situation will become similar to that for the platoon of slow vehicles.

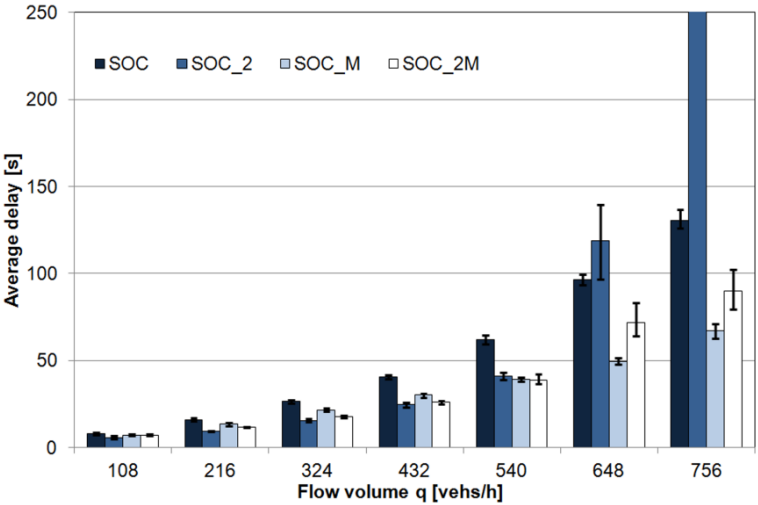

**Fig. 7.** Average delay vs. flow volume (VSN scenario, $f = 20\%$).

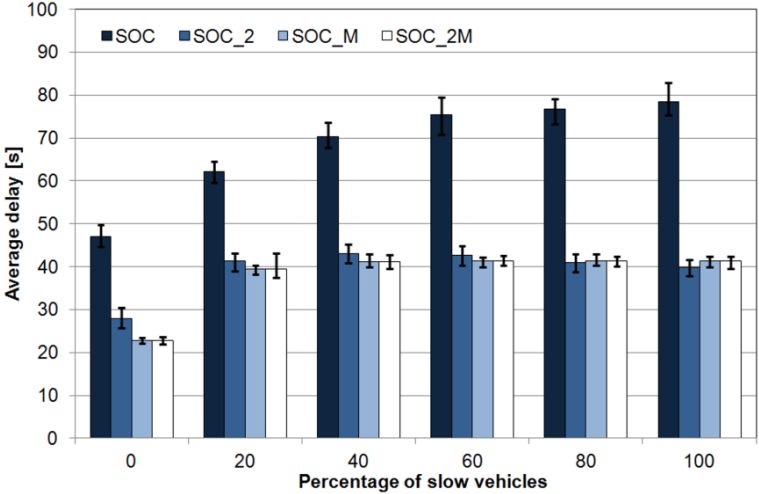

**Fig. 8.** Average delay vs. percentage of slow vehicles (VSN scenario, $q = 540$ vehs/h).

When comparing the results for both simulated scenarios (RVD and VSN) it is evident that the availability of detailed information on vehicle positions in VSN enables better performance of the self-organizing traffic control. It should be noted that for high traffic volumes, the proposed solution (SOC_2M algorithm) used with standard road-side detectors enables better performance than the state-of-the-art SOC algorithm supported by the complex data from VSN. These results are due to the fact that the proposed approach allows the algorithms SOC_2, SOC_M, and SOC_2M to utilize the available traffic information in a more effective way. The interval microscopic model can directly map the positions of vehicles delivered by VSN. In comparison with the macroscopic model used in SOC algorithm, the microscopic model more accurately describes the interactions between individual vehicles and thus better evaluates the influence of the vehicles distribution along a traffic lane on the predicted travel times and delays. As a consequence, the decisions of control agents are more effective.

## 6. CONCLUSION

New algorithms were introduced for decision making by control agents in the self-organizing traffic control system. According to the proposed approach, control decisions are made on the basis of the predicted cost of control actions. The prediction is obtained as a result of traffic simulation in the interval microscopic model, which was developed for this purpose by combining cellular automata and interval arithmetic. This approach allows the traffic control agents to recognize and manage uncertain decisions that can lead to sub-optimal system performance.

Predictions in the proposed system are made on the basis of input data that describe traffic streams at the level of individual vehicles (so-called microscopic level of details). The interval microscopic model enables computationally effective on-line processing of the large input data sets for fast estimation of future vehicle delays. Due to the application of the interval arithmetic, the model is suitable for mapping real-world non-uniform traffic streams, where the parameters of particular vehicles may vary in a given range.

Evaluation of the proposed traffic control strategies was performed in a simulation environment. The traffic simulations were conducted using a model of urban road network based on stochastic cellular automata. The experiments aimed at investigating the performance of the proposed agent decision algorithms in self-organizing traffic control. To this end, average delay was evaluated for various traffic flow volumes and percentages of slow vehicles. Results of the simulations were compared against those obtained by recent algorithms from the literature. This comparison shows that the proposed approach outperforms the state-of-the-art algorithms.

There are two main elements of the proposed approach that have enabled the improvement in performance of the self-organized traffic control. First is the novel interval microscopic traffic model, which allows both the input vehicle parameters and the output predicted delays to be represented by intervals. Thus, instead of average values the control agents can take into account intervals of possible vehicle velocities and queue lengths at intersections. Second element is the procedure proposed for selecting the most effective control action. This procedure selects the control action on the basis of the predicted delay intervals and eliminates uncertain control decisions that may cause increased delays of vehicles due to avoidable signal switching at intersections and the related additional setup time.

The obtained results provide a strong motivation for further research on applications of both the self-organization paradigm and the microscopic models in urban traffic control. The results are encouraging to test the proposed approach in more realistic scenarios, based on examples of real-world road networks. An important issue for future studies is the development of learning methods that would allow the traffic control agents to adapt to changing traffic characteristics. To this end a method will be proposed for tuning the parameters of the interval microscopic model on the basis of the available traffic data in real time. Another interesting direction for further research is to apply the self-organizing traffic control agents in a vehicular ad-hoc network environment. Such agents will work at vehicle nodes to provide dedicated on-board traffic signals for individual vehicles in the road network.

**REFERENCES**

Abdoos, M., Mozayani, N., Bazzan, A. L., 2013. Holonic multi-agent system for traffic signals control. Engineering Applications of Artificial Intelligence 26 (5–6), 1575-1587.

Bernaś, M, 2012. VANETs as a part of weather warning systems, in: Computer Networks. Springer, Berlin Heidelberg, pp. 459-466.

Brockfeld, E., Barlovic, R., Schadschneider, A., Schreckenberg, M., 2001. Optimizing traffic lights in a cellular automaton model for city traffic. Physical Review E, 64(5), 056132.

Burguillo-Rial, J. C., Rodríguez-Hernández, P. S., Montenegro, E. C., and Castiñeira, F. G. 2009. History-based Self-Organizing Traffic Lights. Computing and Informatics, 28(2): 157-168.

Cesme, B., Furth, P. G., 2013. Self-Organizing Control Logic for Oversaturated Arterials. Transportation Research Record: Journal of the Transportation Research Board, 2356(1): 92-99.

Chiu, S., Chand, S., 1993. Self-organizing traffic control via fuzzy logic, in: Proceedings of the 32nd IEEE Conference on Decision and Control. IEEE, pp. 1897-1902.

Cools, S. B., Gershenson, C., D'Hooghe, B., 2013. Self-organizing traffic lights: A realistic simulation in: Advances in Applied Self-Organizing Systems. Springer, London, pp. 45-55.

de Gier, J., Garoni, T. M., Rojas, O., 2011. Traffic flow on realistic road networks with adaptive traffic lights. Journal of Statistical Mechanics: Theory and Experiment, 2011.04: P04008.

Floreano, D., Mattiussi, C., 2008. Bio-inspired artificial intelligence: theories, methods, and technologies. The MIT Press, Cambridge.

Gershenson, C., Heylighen, F., 2003. When can we call a system self-organizing?, in: Advances in Artificial Life. Springer, Berlin Heidelberg, pp. 606-614.

Gershenson, C., 2005. Self-organizing Traffic Lights. Complex Systems, 16, 29–53.

Gershenson, C., Rosenblueth, D. A., 2012. Self-organizing traffic lights at multiple-street intersections. Complexity, 17(4), 23-39.

Gershenson, C., 2013. Living in living cities. Artificial life, 19(3_4), 401-420.

Hamilton, A., Waterson, B., Cherrett, T., Robinson, A., Snell, I., 2013. The evolution of urban traffic control: changing policy and technology. Transportation Planning and Technology, 36(1), 24-43.

Helbing, D., Lämmer, S., Lebacque, J. P., 2005. Self-organized control of irregular or perturbed network traffic, in: Optimal Control and Dynamic Games. Springer US, pp. 239-274.

Ishibuchi, H., Tanaka., H., 1990. Multiobjective programming in optimization of the interval objective function. European Journal of Operational Research, 48, 219-225.

Khamis, M. A., Gomaa, W., 2014. Adaptive multi-objective reinforcement learning with hybrid exploration for traffic signal control based on cooperative multi-agent framework. Engineering Applications of Artificial Intelligence, 29, 134-151.

Lämmer, S., Donner, R., Helbing, D., 2008. Anticipative control of switched queueing systems. The European Physical Journal B, 63(3), 341-347.

Lämmer, S., Helbing, D., 2008. Self-control of traffic lights and vehicle flows in urban road networks. Journal of Statistical Mechanics: Theory and Experiment, 2008(04), P04019.

Maerivoet, S., De Moor, B., 2005. Cellular automata models of road traffic. Physics Reports, 419(1), 1-64.

McKenney, D., White, T., 2013. Distributed and adaptive traffic signal control within a realistic traffic simulation. Engineering Applications of Artificial Intelligence, 26(1), 574-583.

Moore, R. E., 1979. Methods and applications of interval analysis, vol. 2. Siam, Philadelphia.

Nakatsuji, T., Kaku, T., 1991. Development of a self-organizing traffic control system using neural network models. Transportation Research Record, 1324, 137-145.

Pamula, W., 2012. Determination of road traffic parameters based on 3d wavelet representation of an image sequence, in: Computer Vision and Graphics. Springer, Berlin Heidelberg, pp. 541-548.

Płaczek, B., Staniek, M., 2007. Model based vehicle extraction and tracking for road traffic control, in: Computer Recognition Systems 2. Springer, Berlin Heidelberg, pp. 844-851.

Płaczek, B., 2012. Selective data collection in vehicular networks for traffic control applications. Transportation Research Part C: Emerging Technologies, 23, 14-28.

Płaczek, B., 2013. A Traffic Model based on fuzzy cellular automata. Journal of Cellular Automata, 8(3-4), 261-282.

Sekiyama, K. Nakanishi, J. Takagawa, I. Higashi, T., Fukuda, T., 2001. Self-organizing control of urban traffic signal network, in: International Conference on Systems, Man, and Cybernetics, vol. 4, IEEE, pp. 2481-2486.

Suzuki, H., Imura, J. I., Aihara, K, 2013. Chaotic Ising-like dynamics in traffic signals. Scientific reports, 3: 1127.

Szklarski, J, 2010. Cellular automata model of self-organizing traffic control in urban networks. Bulletin of the Polish Academy of Sciences: Technical Sciences, 58(3), 435-441.

Vasirani, M., Ossowski, S., 2011. Learning and coordination for autonomous intersection control. Applied Artificial Intelligence, 25(3), 193-216.

Wang, H., Liu, Y. X., 2013. Self-Organized Traffic Signal Coordinated Control Based on Interactive and Distributed Subarea. Applied Mechanics and Materials, 241, 2031-2037.

Wei, J., Wang, A., Du, N., 2005. Study of self-organizing control of traffic signals in an urban network based on cellular automata. IEEE Transactions on Vehicular Technology, 54(2), 744-748.

Zhang, L., Garoni, T. M., de Gier, J., 2013. A comparative study of Macroscopic Fundamental Diagrams of arterial road networks governed by adaptive traffic signal systems. Transportation Research Part B: Methodological, 49, 1-23.